\documentclass{article}

\usepackage[final]{neurips_2024}

\usepackage{amsthm}
\usepackage{algorithm}
\usepackage{algpseudocode}
\usepackage[utf8]{inputenc} 
\usepackage[T1]{fontenc}    
\usepackage[hidelinks]{hyperref}       
\usepackage{url}            
\usepackage{booktabs}       
\usepackage{amsfonts}       
\usepackage{nicefrac}       
\usepackage{microtype}      
\usepackage{xcolor}
\usepackage{graphicx}
\usepackage{enumitem}
\usepackage{amsmath}
\usepackage{graphicx}  
\usepackage{multirow} 
\usepackage{todonotes}
\usepackage{graphicx}
\usepackage{tikz}
\usepackage[framemethod=default]{mdframed}
\usepackage{parskip} 
\usepackage{mdframed}
\usepackage{xcolor}
\usepackage{graphicx}
\usepackage{subcaption}
\usepackage[most]{tcolorbox}
\usepackage{lmodern}
\usepackage{setspace}
\usepackage{booktabs}
\usepackage{float}
\usepackage{adjustbox}
\usepackage{wrapfig}
\usepackage{soul}
\usepackage[T1]{fontenc}

\usepackage{microtype}
\usepackage[normalem]{ulem}
\usepackage{todonotes}

\newcommand{\gcite}{{\tt{G-Cite}}}
\newcommand{\pcite}{{\tt{P-Cite}}}

\title{Generation-Time vs. Post-hoc Citation: A Holistic Evaluation of LLM Attribution}

\author{%
  Yash Saxena \\
  UMBC\\
  Baltimore, Maryland, USA \\
  \texttt{ysaxena1@umbc.edu} \\
  \And
  Raviteja Bommireddy\\
  IIITDM  \\
  Kancheepuram, Chennai, India\\
  \texttt{cs23b2011@iiitdm.ac.in} \\
  \AND
  Ankur Padia\\
  UMBC \\
  Baltimore, Maryland, USA \\
  \texttt{pankur1@umbc.edu} \\
  \And
  Manas Gaur \\
  UMBC \\
  Baltimore, Maryland, USA \\
  \texttt{manas@umbc.edu} \\
}

\definecolor{codebg}{HTML}{F8F9FA}
\definecolor{promptborder}{HTML}{4EA72E}
\definecolor{highlightyellow}{HTML}{FBC02D}
\begin{document}

\maketitle

\begin{abstract}
Trustworthy Large Language Models (LLMs) must cite human-verifiable sources in high-stakes domains such as healthcare, law, and scientific research, where even small errors can have severe consequences. Practitioners and researchers face a choice: let models generate citations during decoding, or let models draft answers first and then attach appropriate citations. To clarify this choice, we introduce two paradigms: \emph{Generation-Time Citation} (\gcite{}), which produces the answer and citations in one pass, and \emph{Post-hoc Citation} (\pcite{}), which adds or verifies citations after drafting. We conduct a comprehensive evaluation from zero-shot to advanced retrieval-augmented methods across four popular attribution datasets, and provide evidence-based recommendations that weigh trade-offs across use cases. Our results show a consistent trade-off between coverage and citation correctness, with retrieval as the main driver of attribution quality in both paradigms. \pcite{} methods achieve high coverage with competitive correctness and moderate latency, whereas \gcite{} methods prioritize precision at the cost of coverage and speed. We recommend a retrieval-centric, \pcite{}-first approach for high-stakes applications, reserving \gcite{} for precision-critical settings such as strict claim verification.
Our codes and human evaluation results are available at \url{https://anonymous.4open.science/r/Citation_Paradigms-BBB5/}

\end{abstract}

\section{Introduction}
\label{Introduction}

Just as humans cite sources to demonstrate credibility in their communication, LLM-based AI systems must provide attribution to build trust in their outputs \citep{phukan-etal-2024-peering, 10.1145/3555803}. Trustworthy AI has become a national priority following the White House's Executive Order on AI \citep{whitehousePreventingWoke}, and this will become increasingly critical as models scale. Researchers and practitioners working in high-stakes applications need confidence that their LLM-based  AI tools are reliable and transparent \citep{leyliabadi2025conceptualframeworkaibaseddecision, kowald2024establishing}. To contextualize, consider the legal domain where summarizing lengthy documents is routine. In such settings, it is essential to attribute each generated sentence to its corresponding source text to ensure reliability and transparency \citep{batista2025think}. Viewing attribution as a practical pathway to trustworthiness, we categorize existing research into two fundamental paradigms \gcite{} and \pcite{}, providing researchers and practitioners with actionable choices for developing attribution-capable LLMs.



\emph{Generation-Time Citation} (\gcite{}) creates the text and citation markers together in a single step. \emph{Post-hoc Citation} (\pcite{}) works differently; it first creates a draft, then adds or checks citations in a separate step. The key difference between these approaches is the timing of citations. Besides timings, these approaches differ in how they work technically. \gcite{} makes citation choices during the normal left-to-right text generation process. It decides \textit{locally} based on what has been written so far and any retrieved evidence. In contrast, \pcite{} separates citation from text generation entirely. It runs a second pass to examine the complete draft and available evidence, adding or verifying citations throughout the text. 


\begin{wraptable}{r}{0.5\textwidth}
\centering
\vspace{-0.4cm}
\footnotesize
\setlength{\tabcolsep}{6pt}
\resizebox{0.5\textwidth}{!}{
\begin{tabular}{l l l l r}
\toprule
\textbf{Dataset} & \textbf{Citation granularity} & \textbf{Nativeness} & \textbf{Instances (\#)} \\
\midrule
\textbf{ALCE} 
  & Doc, Sent 
  & \gcite{} 
  & 3{,}000 \\
\textbf{LongCite} 
  & Sent 
  & \gcite{} 
  & 1{,}000 \\
\textbf{REASONS} 
  & Doc, Sent 
  & \pcite{} 
  & 12{,}723 \\
\textbf{FEVER} 
  & Sent, Claim 
  & \pcite{} 
  & 185{,}445 \\
\bottomrule
\end{tabular}}
\caption{\scriptsize \textbf{Details of Datasets used in the Experiments.} FEVER focuses on factual verification of claims in health, law, and other domains. REASONS is focused on scientific research, and LongCite/ALCE are open domain QA with long and short context, respectively. ``Doc'', ``Sent'', ``Claim'' denote if the citation is at document-level, sentence-level, or Claim-level.}
\label{tab:datasets}
\vspace{-0.3cm}
\end{wraptable}

As of now, there is no principal and systematic study to compare these two paradigms to determine their limitations and capabilities. Each method is designed differently and evaluated using a different evaluation metric on different datasets. These gap makes it difficult for researchers and practitioners to choose the right method or design better attribution systems.  To address this problem, our paper provides a rigorous empirical comparison of both paradigms using standard datasets and evaluation metrics across diverse categories of methods. 


\begin{wrapfigure}[25]{r}{6.8cm}
    \centering
    \vspace{-0.5cm}
    \includegraphics[scale=0.37]{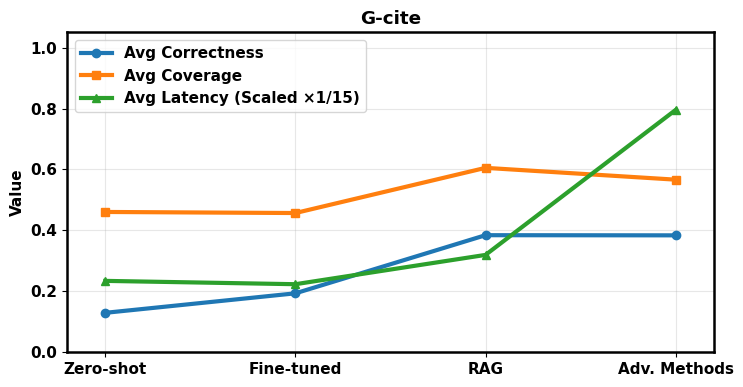}
    \includegraphics[scale=0.37]{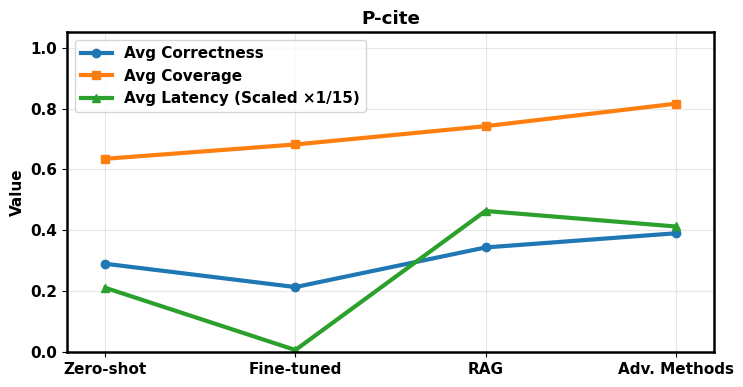}
    \scriptsize
    \caption{\footnotesize \textbf{Citation Quality Trends.} Average citation correctness, entailed coverage, and latency across categories (Zero-shot, Fine-tuned, RAG, Advanced) for the \gcite{} and \pcite{} paradigms, averaged over all datasets.}
\label{fig:generation_cuad_512}
\end{wrapfigure}
We benchmark the state-of-the-art methods from both paradigms. Because several existing datasets are tailored to a single paradigm, we adapt them to enable fair, cross-paradigm comparison. More details on the datasets are available in \autoref{tab:datasets}. We consider four types of methods: Zero-shot, Fine-tuned, Retrieval Augmented Generation (RAG), and two recent approaches from \gcite{} and \pcite{} as Advanced methods. Here, zero-shot and RAG act as common baselines. For Advanced and Fine-tuned methods, within \gcite{}, we use CoT Citation \citep{Ji_Liu_Du_Ng_2024} and LongCite (8b) \citep{zhang-etal-2025-longcite} and within \pcite{}, we evaluate CiteBART \citep{celik2025citebartlearninggeneratecitations} and Citation-Enhanced Generation (CEG) \citep{li-etal-2024-citation}. 
We assess performance using established quantitative attribution metrics: Citation Correctness, Precision, Recall, Coverage, and Latency. These metrics are widely used in prior work and enable consistent comparison across methods. We also conduct a human evaluation with n=100 instances per method-dataset pair, providing 80\% statistical power to detect medium effect sizes (Cohen's d $\geq$ 0.5) at $\alpha$ = 0.05 significance level, for more details, see \autoref{Experimentation}.

\begin{figure*}
    \includegraphics[width=\linewidth]{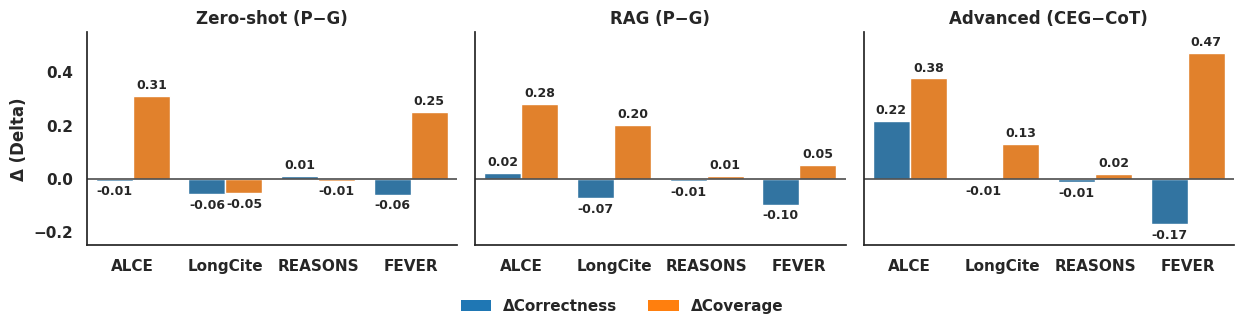}
    \caption{\footnotesize \textbf{Coverage and correctness deltas (\pcite{} - \gcite{}) across datasets.} 
    Positive values indicate \pcite{} outperforms \gcite{} on the metric; negative values indicate otherwise.}
    \label{fig:delta_fig}
\end{figure*}

\noindent \textbf{Findings:} By analyzing results shown in \autoref{fig:generation_cuad_512} and \autoref{fig:delta_fig}, we establishes four primary insights: (1) Retrieval augmentation is fundamental as it provides the largest gains in both citation correctness and coverage regardless of paradigm choice; (2) \pcite{} methods achieve higher coverage with competitive citation correctness as they include more ground-truth citations while maintaining reasonable citation precision and citation recall balance; (3) Advanced methods enable targeted optimization as they allow practitioners to adjust the citation precision and coverage with latency costs; and (4) Organizations should view fine-tuning as an optimization enhancement rather than a replacement strategy, while domain-specific models can improve efficiency and task alignment, retrieval-augmented approaches remain essential to maintain the citation correctness and coverage standards when information accuracy is non-negotiable.

\section{Related Work}

Recent research has focused heavily on enhancing the ability of LLMs to generate correct source citations. These efforts can be broadly categorized into two groups based on the underlying paradigm they follow.

\subsection{Generation-time Citation (\gcite)}
\label{G-Cite}
\textbf{Methods:} Methods in this category, which we term G-Cite, generate citations concurrently with the text. Foundational work in this area includes benchmarks like ALCE and CiteBench \citep{funkquist-etal-2023-citebench}, which provide datasets and prompting-based baselines. Building on these, other prompting-based approaches, such as Learning to Plan \citep{fierro-etal-2024-learning}, further refine the attribution capabilities of LLMs. In contrast to prompting, methods like FRONT \citep{huang-etal-2024-training}, LongCite \citep{zhang-etal-2025-longcite}, and Self-Cite \citep{chuang2025selfcite} aim to improve source citation by fine-tuning the models themselves. Additionally, methods such as ReCLAIM \citep{xia-etal-2025-ground} and Chain-of-Thought (CoT) Citation \citep{Ji_Liu_Du_Ng_2024} build on RAG systems to enhance source citation capabilities. Other works focus on specific aspects, such as the granularity of the generated citations \citep{li-etal-2024-cited}.

\textbf{Datasets:} In open domain short-context datasets, where the documents containing the citation to the correct answer are relatively short, include ASQA \citep{stelmakh-etal-2022-asqa}, QAMPARI \citep{amouyal-etal-2023-qampari}, and ALCE. For long-context datasets, where the documents containing the citation are relatively long, there are popular datasets such as ELI5 \citep{fan-etal-2019-eli5} and LongBench-Cite. 

\subsection{Post-hoc Citation (\pcite)}
\label{P-Cite}

\textbf{Methods:} The second category, P-Cite, includes methods that generate citations for a pre-existing text. A prominent example is the REASONS benchmark \citep{saxena2025attributionscientificliteraturenew}, which provides a dataset and Retrieval-Augmented Generation (RAG) baselines for sentence-level citation. Other methods also leverage RAG, using iterative approaches \citep{li-etal-2024-citation} or adding a verification step for each claim, as seen in RARR. Another recent approach that builds on top of RAG systems is CiteFix \citep{maheshwari-etal-2025-citefix}, which uses a two-step citation correction method to achieve better citation quality. Beyond standard RAG, approaches in this paradigm include fine-tuning with models like CiteBART \citep{celik2025citebartlearninggeneratecitations} and using hierarchical attention mechanisms as in HAtten \citep{10.1007/978-3-030-99736-6_19}. Research in P-Cite also addresses related sub-tasks, such as recommending missed citations \citep{long-etal-2024-recommending} or determining if a sentence requires a citation at all \citep{batista2025thinkattributeimprovingperformance}. 

\textbf{Datasets:} Open-domain, claim-level datasets such as FEVER,  fall into this category. In addition, scientific datasets such as RefSeer \citep{10.5555/2740769.2740832}, PeerRead (FullTextPeerRead) \citep{kang-etal-2018-dataset}, ACL-ARC \citep{bird-etal-2008-acl}, SciFact (+SciFact-Open) \citep{wadden-etal-2020-fact}, and REASONS are \pcite{} native. Legal datasets such as Bar Exam QA, and Housing Statutes QA \citep{10.1145/3709025.3712219} also belong to this category.

\section{Experimentation}
\label{Experimentation}


We benchmark state-of-the-art methods from both paradigms, \gcite{} and \pcite{}, eight in total (four per paradigm). For a fair evaluation, we use \textit{LLaMa-3.1-8B-Instruct} for all methods except for CiteBART which makes use of the BART model. 



\textbf{Datasets.} \autoref{tab:datasets} shows the datasets that span open-domain QA, scientific citation, and fact verification; long and short context; \gcite{}-native and \pcite{}-native settings; and sentence- and document-level granularity. 
For fair comparison, we adapt each dataset to also support the non-native paradigm. For ALCE \citep{gao-etal-2023-enabling} and LongBench-Cite (both \gcite{}-native, where the model outputs an answer with inline markers), we create a citation-free draft answer and then let \pcite{} attach inline citations. For REASONS (\pcite{}-native), we evaluate \gcite{} by providing a constrained candidate pool (titles) and prompting the model to rewrite the target sentence with an inline citation. For FEVER \citep{thorne-etal-2018-fever}, we treat each claim as a “needs citation” unit: \gcite{} generates the statement with an inline citation, while \pcite{} attaches a citation to the fixed statement.

\textbf{Methods.} We evaluate \gcite{} and \pcite{} using a systematic categorization of four method types: (i) Zero-shot prompting, (ii) Fine-tuned models, (iii) Retrieval-augmented methods, and (iv) Advanced hybrid techniques. Zero-shot prompting was used across both the \gcite{} and \pcite{} paragraphs, serving as a fundamental baseline without task-specific training or fine-tuning. Implementation details and zero-shot prompts with examples are in \autoref{implementation_details}. \textit{Fine-tuned models} represent specialized architectures trained on domain-specific datasets, such as LongCite-8B for citation generation and CiteBART for paper citations, both trained using supervised learning on ALCE and arXiv datasets. Retrieval-augmented methods combine retrieval systems with generative models to enhance citation accuracy and coverage. We considered retrieval-based methods in both \gcite{} and \pcite{}. Advanced hybrid techniques integrate multiple methodological components: CoT Citation combines evidence retrieval with chain-of-thought prompting and includes an evidence-insurance step for comprehensive citation coverage, while CEG employs an iterative approach that retrieves relevant information before systematically attaching and verifying citations. 

\begin{wrapfigure}[18]{r}{5cm}
    \centering
\includegraphics[width=5cm]{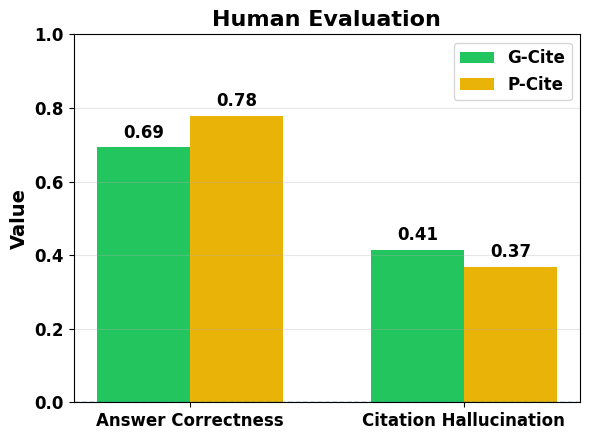}
    \scriptsize
    \caption{\scriptsize
    \textbf{Human Evaluation Results.} We report \emph{Answer Correctness} ($\uparrow$), and \emph{Citation Hallucination} ($\downarrow$), values are averaged over all datasets and methods within each paradigm (\gcite{} and \pcite{}).\pcite{} based methods tend to provide more correct answers with lesser hallucination.}
\label{fig:Human_Eval}
\end{wrapfigure}

\textbf{Metrics and Human Evaluation.} We use standard citation metrics for evaluation. Specifically, we use five standard metrics: (i) Citation Precision to measure the fraction of correct citations among those produced; ii) Citation Recall to measure the fraction of ground-truth citations that are retrieved;(iii) Citation Correctness to measure the harmonic mean of precision and recall \citep{aly-etal-2024-learning};
(iv) Coverage to measure the proportion of ground-truth citations present in the generated response \citep{aly-etal-2024-learning}; (v) Latency to measure the average time (in seconds) taken by each method per dataset instance. We conducted a human assessment using two expert annotators from the university library with experience in AI-assisted citation verification. We evaluated 100 instances per method-dataset pair across two critical quality measures for each generated response, achieving $\kappa$ = 0.873 inter-annotator agreement.
\textit{Answer Correctness} is a strict metric that allows human evaluators to verify whether the provided evidence actually supports each claim made in the generated text. 
We assign a score of 1 when all claims are properly backed by their cited sources, and 0 when any claims lack adequate support. \textit{Citation Hallucination} allows human evaluators to check whether each citation corresponds to a real source from the reference dataset. Humans assign a score of 1 when citations are fabricated or point to sources outside the ground truth collection, and 0 when all citations are legitimate and verifiable. 




\section{Results and Analysis}




Our evaluation reveals distinct performance characteristics between the two citation paradigms across all datasets. \pcite{} consistently achieves higher coverage while maintaining competitive citation correctness compared to \gcite{}. For practitioners deploying LLMs in information-seeking applications, where users need comprehensive source attribution to verify claims across multiple documents, \pcite{} methods provide a critical advantage by ensuring broader citation coverage without compromising accuracy. Further, as shown in \autoref{fig:Human_Eval}, the human evaluation reinforces these findings: averaged over datasets and methods, \pcite{} shows higher answer correctness than \gcite{} (78\% vs. 69\%) and lower citation hallucination (37\% vs. 41\%). See Appendix \ref{results_tables} for detailed results.

\textbf{Finding 1:} On ALCE, the advanced \pcite{} achieve 75\% coverage with 42\% correctness, substantially outperforming the advanced \gcite{} which reaches 37\% coverage and 21\% correctness. Similarly, on LongBench-Cite, \pcite{} attains 78\% coverage with 12\% correctness, while \gcite{} achieves 65\% coverage with 12\% correctness. This shows that in complex information synthesis tasks, such as research summaries, technical reports, or knowledge synthesis, practitioners can expect \pcite{} methods to provide citations for approximately twice as many relevant sources as \gcite{} methods, dramatically improving the verifiability and trustworthiness of generated content.

\textbf{Finding 2:} On the scientific REASONS dataset, both paradigms achieve comparable correctness (\pcite{} 26\% vs. \gcite{} 27\%) and near-ceiling coverage (\pcite{} 99\% vs. \gcite{}: 97\%). On the FEVER dataset, \gcite{} achieves the highest precision and correctness (94\%) but limited citation coverage (27\%), while \pcite{} provides a more balanced profile with high coverage (74\%) and strong correctness (75\%). This shows that in scientific literature tasks, both P-cite and G-cite methods perform similarly well. However, for fact verification tasks, which are important for legal and policy-driven research, P-cite methods excel when you need comprehensive evidence from multiple sources, while G-cite methods work better when you need extremely precise validation of individual claims.

\textbf{Finding 3:} Retrieval augmentation emerges as the primary driver of citation accuracy. The transition from zero-shot to RAG yields the most substantial and consistent improvements across both paradigms and all datasets. On FEVER, \gcite{} correctness improves by approximately 50 percentage points (from 27\% to 77\%), while on LongBench-Cite, coverage increases by approximately 47 percentage points (from 11\% to 58\% for \gcite{}). Our results clearly suggest that organizations deploying LLMs for information-critical applications should prioritize investment in retrieval infrastructure as the foundational requirement to gain access to relevant, high-quality source material for LLM-based applications. 

\textbf{Finding 4:} Advanced methods built on retrieval foundations adjust the citation coverage and latency trade-off. \pcite{} delivers high coverage and correctness with moderate latency costs, whereas \gcite{} delivers better performance but substantially increases latency. Practitioners must balance operational efficiency with accuracy, where \pcite{} offers a practical solution and \gcite{} could be used for verification purposes. Fine-tuned models provide incremental improvements but cannot replace retrieval for maintaining content accuracy. Human evaluators, representing end users who rely on generated content for critical decisions, consistently rate \pcite{} outputs as more accurate and trustworthy.

\section{Conclusion}
In this work, we empirically evaluated \gcite{} and \pcite{}, highlighting their respective capabilities and limitations. Our findings show that retrieval-based attribution is fundamental regardless of paradigm. Using common metrics and datasets, we demonstrate that \pcite{} is better suited for high-stakes applications due to higher factuality, while \gcite{} is preferable in precision-critical settings. To facilitate future research, we release our code and human evaluations for reproducibility.

\section{Acknowledgment}

We thank the human annotators who contributed to this study for their valuable efforts. We also thank the anonymous reviewers for their constructive feedback and suggestions, which helped improve the quality of this work. We gratefully acknowledge support from the UMBC Faculty Start-up, Cybersecurity Leadership-Exploratory Grant and the USISTEF Award. The opinions, conclusions, and recommendations expressed here are solely those of the authors and do not necessarily reflect the views of USISTEF or UMBC.

\bibliographystyle{unsrtnat}
\bibliography{main}

\newpage
\appendix

\section{Appendix}
\label{Appendix}

\subsection{Additional Implementation Details}
\label{implementation_details}

We implement all citation variants in PyTorch with Hugging Face Transformers. ALCE inputs (both `qa pairs` blocks and flat items) are normalized, then sentence‐split with a lightweight regex and numbered to form the Context. \textbf{Answer Generation} we keep `MAX\_NEW\_TOKENS' as 256, decoding uses `temperature=0.2', `top\_p=0.95'. We parse citations with a strict regex, remap packed indices back to original sentences, and rebuild a clean \textbf{References} block (title + sentence). For RAG P/G-cite runs we add SBERT bi-encoder retrieval and a cross-encoder reranker before prompting; zero-shot variants skip retrieval.

\begin{tcolorbox}[
  colback=codebg,
  colframe=promptborder,
  title=\textbf{\gcite{} Zero-shot Prompt with Example},
  fonttitle=\bfseries\scriptsize,
  coltitle=white,
  arc=3pt,
  left=6pt,
  right=6pt,
  top=6pt,
  bottom=6pt,
  width=\linewidth,
  fontupper=\scriptsize,
  enhanced,
]
\textbf{Instructions:} Answer the QUESTION using ONLY the numbered Context sentences. 
After each answer sentence (or clause), append a citation tag exactly like <cite>[i]</cite> or <cite>[i-j]</cite> that points to the supporting Context index(es).

\begin{itemize}[leftmargin=10pt,itemsep=0em,topsep=0pt]
    \item Every factual claim must have at least one <cite>[i]</cite> tag.
    \item After the answer, output a References block that lists only the indices you cited, one per line as: [k] <Context title> — sent local\_sent\_index: verbatim Context sentence.
    \item Do NOT add explanations; return only the answer and the References block.
\end{itemize}


\textbf{Example Query:} \textit{Who is the original artist of “The Sound of Silence”?}

\textbf{Example Response:}
\begin{flushleft}
\scriptsize
\texttt{The original artist of “The Sound of Silence” is the American duo Simon \& Garfunkel <cite>[1]</cite>.}
\\
\texttt{References:
[1] (The Sound of Silence) "The Sound of Silence" is a song by the American music duo Simon \& Garfunkel.} \\
\end{flushleft}
\end{tcolorbox}

\begin{tcolorbox}[
  colback=codebg,
  colframe=highlightyellow,
  title=\textbf{\pcite{} Zero-shot Prompt with Example},
  fonttitle=\bfseries\scriptsize,
  coltitle=white,
  arc=3pt,
  left=6pt,
  right=6pt,
  top=6pt,
  bottom=6pt,
  width=\linewidth,
  fontupper=\scriptsize,
  enhanced,
]
\textbf{Instructions:} You will answer the QUESTION and then insert inline numeric citations [k] to that answer after the facts you assert.

\begin{itemize}[leftmargin=10pt,itemsep=0em,topsep=0pt]
    \item Insert [k] markers immediately after the clause(s) they support.
    \item The References section must contain ONLY markers that appear in the sentence.
    \item Do NOT include explanations; return only the answer and the References block.
\end{itemize}


\textbf{Example Query:} \textit{When was the first iPhone released?}

\vspace{4pt}
\textbf{Example Response:}
\begin{flushleft}
\scriptsize
\texttt{The first iPhone was released on June 29, 2007 [1].}
\\
\texttt{References:
[1] https://en.wikipedia.org/wiki/IPhone\_(1st\_generation)} \\
\end{flushleft}
\end{tcolorbox}

\subsection{Results Tables}
\label{results_tables}

\begin{table*}[!htbp]
\centering
\scriptsize
\setlength{\tabcolsep}{4.5pt}

\resizebox{\textwidth}{!}{
\begin{tabular}{l l r r r r r}
\toprule
\textbf{Dataset} & \textbf{Method (Paradigm)} & \textbf{Cit. Corr. $\uparrow$} & \textbf{Cit. Prec. $\uparrow$} & \textbf{Cit. Rec. $\uparrow$} & \textbf{Cov. $\uparrow$} & \textbf{Latency $\downarrow$} \\
\midrule
\multirow{10}{*}{\textbf{ALCE}}
  & \textit{Zero-shot (G)}                     & 0.130 & 0.156 & 0.111 & 0.274 & 2.925 \\
  & \textit{RAG (G)}                            & 0.319 & 0.422 & 0.257 & 0.340 & 6.513 \\
  & CoT Citation (G)                            & 0.205 & 0.239 & 0.180 & 0.372 & 17.237 \\
  & LongCite (8B) (G)                           & 0.253 & 0.271 & 0.236 & 0.282 & 3.531 \\
  & \textit{Zero-shot (P)}                      & 0.881 & 1.000 & 0.787 & 0.784 & 4.743 \\
  & \textit{RAG (P)}                            & 0.340 & 0.441 & 0.277 & 0.620 & 9.059 \\
  & CiteBART (P)                                & --- & --- & --- & --- & --- \\
  & CEG (P)                                     & 0.422 & 0.626 & 0.318 & 0.748 & 6.077 \\
\cmidrule(lr){1-7}
\multirow{10}{*}{\textbf{LongBench-Cite}}
  & \textit{Zero-shot (G)}                      & 0.099 & 0.127 & 0.081 & 0.112 & 3.211 \\
  & \textit{RAG (G)}                            & 0.167 & 0.163 & 0.171 & 0.577 & 5.533 \\
  & CoT Citation (G)                            & 0.121 & 0.155 & 0.101 & 0.652 & 17.476 \\
  & LongCite (8B) (G)                           & 0.134 & 0.173 & 0.097 & 0.632 & 3.171 \\
  & \textit{Zero-shot (P)}                      & 0.040 & 0.569 & 0.021 & 0.058 & 3.407 \\
  & \textit{RAG (P)}                            & 0.093 & 0.098 & 0.088 & 0.780 & 5.842 \\
  & CiteBART (P)                                & --- & --- & --- & --- & --- \\
  & CEG (P)                                     & 0.115 & 0.435 & 0.066 & 0.782 & 9.694 \\
\bottomrule
\end{tabular}}
\caption{Open-domain results (ALCE and LongBench-Cite). Metrics: Citation Correctness (Corr.), Citation Precision/Recall (Prec./Rec.), Coverage (Cov.), and Latency (s). A dash (---) marks method–dataset pairs we did \emph{not} execute due to implementation constraints (e.g., domain-locked models, unavailable code/weights).}
\label{tab:results_open}
\end{table*}

\begin{table*}[!htbp]
\centering
\scriptsize
\setlength{\tabcolsep}{4.5pt}

\begin{tabular}{l l r r r r r}
\toprule
\textbf{Dataset} & \textbf{Method (Paradigm)} & \textbf{Cit. Corr. $\uparrow$} & \textbf{Cit. Prec. $\uparrow$} & \textbf{Cit. Rec. $\uparrow$} & \textbf{Cov. $\uparrow$} & \textbf{Latency $\downarrow$} \\
\midrule
\multirow{10}{*}{\textbf{REASONS}}
  & \textit{Zero-shot (G)}           & 0.017 & 0.015 & 0.020 & 0.954 & 3.719 \\
  & \textit{RAG (G)}                  & 0.282 & 0.268 & 0.298 & 0.802 & 4.111 \\
  & CoT Citation (G)                  & 0.272 & 0.244 & 0.306 & 0.970 & 9.567 \\
  & LongCite (8B) (G)                 & ---   & ---   & ---   & ---   & ---   \\
  & \textit{Zero-shot (P)}            & 0.029 & 0.027 & 0.032 & 0.946 & 3.535 \\
  & \textit{RAG (P)}                  & 0.272 & 0.269 & 0.276 & 0.814 & 10.083 \\
  & CiteBART (P)                      & 0.114 & 0.139 & 0.097 & 0.682 & ---   \\
  & CEG (P)                           & 0.259 & 0.241 & 0.280 & 0.989 & 6.628 \\
\bottomrule
\end{tabular}
\caption{Scientific-domain results (REASONS). Metrics: Citation Correctness (Corr.), Citation Precision/Recall (Prec./Rec.), Coverage (Cov.), and Latency (s). G-cite runs use a constrained “rewrite with [k]” adapter over the dataset’s candidate pool.}
\label{tab:results_reasons}
\end{table*}

\begin{table*}[!htbp]
\centering
\scriptsize
\setlength{\tabcolsep}{4.5pt}

\begin{tabular}{l l r r r r r}
\toprule
\textbf{Dataset} & \textbf{Method (Paradigm)} & \textbf{Cit. Corr. $\uparrow$} & \textbf{Cit. Prec. $\uparrow$} & \textbf{Cit. Rec. $\uparrow$} & \textbf{Cov. $\uparrow$} & \textbf{Latency $\downarrow$} \\
\midrule
\multirow{10}{*}{\textbf{FEVER}}
  & \textit{Zero-shot (G)}  & 0.272 & 0.287 & 0.258 & 0.502 & 4.204 \\
  & \textit{RAG (G)}        & 0.769 & 0.781 & 0.757 & 0.702 & 3.017 \\
  & CoT Citation (G)        & 0.937 & 1.000 & 0.881 & 0.272 & 3.439 \\
  & LongCite (8B) (G)     & ---   & ---   & ---   & ---   & ---   \\
  & \textit{Zero-shot (P)}  & 0.212 & 0.344 & 0.153 & 0.752 & 1.011 \\
  & \textit{RAG (P)}        & 0.671 & 0.717 & 0.630 & 0.754 & 2.840 \\
  & CiteBART (P)          & ---   & ---   & ---   & ---   & ---   \\
  & CEG (P)                 & 0.766 & 0.827 & 0.713 & 0.744 & 2.370 \\
\bottomrule
\end{tabular}
\caption{Fact Verification-control results (FEVER). Metrics: Citation Correctness (Corr.), Citation Precision/Recall (Prec./Rec.), Coverage (Cov.), and Latency (s). FEVER is used to calibrate the evidence-agreement judge and to report supported-claim rate.}
\label{tab:results_fever}
\end{table*}

\end{document}